\documentclass[nonacm,sigconf]{acmart}
\usepackage{xcolor}
\usepackage[usenames,dvipsnames]{pstricks}
\usepackage{xcolor}
\usepackage{epsfig}
\usepackage{url}
\usepackage{pst-grad} 
\usepackage{pst-plot} 
\usepackage[space]{grffile} 
\usepackage{etoolbox}
\makeatletter
\usepackage{hhline}
\patchcmd\Gread@eps{\@inputcheck#1 }{\@inputcheck"#1"\relax}{}{}
\makeatother
\usepackage{graphicx}
\usepackage[utf8]{inputenc}
\usepackage{amsmath}
\usepackage{booktabs}
\usepackage{algorithm}
\usepackage{algpseudocode}
\usepackage{listings}
\usepackage{enumitem}
\usepackage{tabu}

\newcommand{\voc}[2]{\texttt{#1:\allowbreak #2}}

\renewcommand\footnotetextcopyrightpermission[1]{}
\fancypagestyle{specialfooter}{%
  \fancyhf{}
  
  \fancyfoot[L]{©2020 Copyright held by the authors. This is the author's version of the work. It is posted here for your personal use. Not for redistribution. The definitive version was published in the Proceedings of the 29th ACM International Conference on Information and Knowledge Management, \url{https://doi.org/10.1145/3340531.3412760}.}
}

\begin{document}

\title{Event-QA: A Dataset for Event-Centric Question Answering over Knowledge Graphs}

\author{Tarcísio Souza Costa, Simon Gottschalk, Elena Demidova}
\affiliation{%
  \institution{L3S Research Center, Leibniz Universit\"at Hannover, Hannover, Germany}
}
\email{{souza, gottschalk, demidova}@L3S.de}

\begin{abstract}
Semantic Question Answering (QA) is a crucial technology to facilitate intuitive user access to semantic information stored in knowledge graphs. Whereas most of the existing QA systems and datasets focus on entity-centric questions, very little is known about these systems' performance in the context of events. As new event-centric knowledge graphs emerge, datasets for such questions gain importance. In this paper, we present the Event-QA dataset for answering event-centric questions over knowledge graphs. Event-QA contains 1000 semantic queries and the corresponding English, German and Portuguese verbalizations for EventKG - an event-centric knowledge graph with more than 970 thousand events.
\end{abstract}

\keywords{Question Answering, Events, Knowledge Graphs, Dataset}

\maketitle
\thispagestyle{empty}
\thispagestyle{specialfooter}

\noindent
\textbf{Resource type:} Dataset \\
\textbf{Resource DOI: 10.5281/zenodo.3568387} \\
\textbf{Permanent URL:} \url{http://eventcqa.l3s.uni-hannover.de} \\

\section{Introduction}
\label{sec:introduction}

Knowledge graphs (KGs) with popular examples including DBpedia \cite{dbpedia-swj}, Wikidata \cite{Vrandecic:2012}, YAGO \cite{RebeleSHBKW16} and EventKG \cite{Gottschalk:2018, GottschalkD19} have recently evolved as an important source of semantic information on the Web. 
Semantic Question Answering (QA) is a key technology
to facilitate intuitive access to knowledge graphs for end-users
via natural language (NL) interfaces. 
QA systems automatically translate a user question expressed in a natural language into the SPARQL query language to query knowledge graphs \cite{TrivediMDL17, lcquad2}.
In recent years, researchers have developed a large variety of QA approaches \cite{HoffnerWMULN17}.

QA datasets play an essential role in the development and evaluation of QA systems \cite{lcquad2, UsbeckGN018}.
The active development of QA datasets supports the advancement of QA technology, for example, through the QALD (Question Answering over Linked Data) initiative\footnote{Question Answering over Linked Data: \url{http://qald.aksw.org}}. 
As popular knowledge graphs are mostly entity-centric, they do not sufficiently cover events and temporal relations \cite{Gottschalk:2018}. As a consequence, existing QA datasets, with recent examples including LC-QuAD \cite{TrivediMDL17}, LC-QuAD 2.0 \cite{lcquad2} and the QALD challenges, mainly focus on entity-centric queries.
More recently, event-centric knowledge graphs such as EventKG \cite{Gottschalk:2018} that includes historical and important contemporary events as well as knowledge graphs that contain daily happenings extracted from the news (e.g., \cite{Leetaru:2013, RospocherEVFARS16}) evolved. However, there is a lack of QA datasets dedicated to event-centric questions and only a few datasets (e.g., TempQuestions \cite{TempQuestions} and the dataset proposed by Saquete et al. \cite{SaqueteGMML09}) focus on temporal expressions.
This way, development and evaluation of QA systems concerning the event-centric questions are currently not adequately supported.

In this paper, we introduce novel \textit{Event-QA} (Event-Centric Question Answering Dataset) dataset for training and evaluating QA systems on complex event-centric questions over knowledge graphs. 
With complex, we mean that the intended SPARQL queries consist of more than one triple pattern, similar to \cite{TrivediMDL17}. %
The aims of \textit{Event-QA} are to: 
1) Provide a representative selection of relations involving events in the knowledge graph, to ensure the diversity of the resulting event-centric semantic queries. To achieve this goal, we approach the query generation via a random walk through the knowledge graph, starting from randomly selected relations.
2) Ensure the quality of the natural language questions. 
To this extent, the resulting queries are manually translated into natural language expressions in English, Portuguese and German.
To the best of our knowledge, this is the first QA dataset focused on event-centric questions so far, and the only QA dataset that targets EventKG.

The contributions of this paper are as follows: 
\begin{itemize}
\item We propose an approach for an automatic generation of semantic queries for an event-centric QA dataset from a knowledge graph. 
\item We present the \textit{Event-QA} dataset containing 3000 event-centric natural language questions (i.e. 1000 in each language) with the corresponding SPARQL interpretations for the EventKG knowledge graph. These questions exhibit a variety of SPARQL queries and their verbalizations.
To facilitate an easier integration with the existing QA pipelines, we also provide a translation of the SPARQL queries in \textit{Event-QA} to DBpedia, where possible.
\item We provide an open source, extensible framework for automatic dataset generation to facilitate dataset maintenance and updates.  
\end{itemize}

The remainder of the paper is organised as follows: In Section \ref{sec:relevance}, we discuss the motivation for an event-centric QA dataset. Then, in Section \ref{sec:problem}, we present the problem of the event-centric QA dataset generation. In Section \ref{sec:approach}, we describe the corresponding approach to generate an event-centric QA dataset from a knowledge graph. Following that, in Section \ref{sec:implementation}, we provide a brief overview of EventKG and the application of the proposed approach to this knowledge graph. Section \ref{sec:evaluation} discusses the characteristics of the resulting \textit{Event-QA} dataset, and Section \ref{sec:availability} summarises the availability aspects. Then, Section \ref{sec:related} gives an overview of the related work. Finally, Section \ref{sec:conclusion} provides a conclusion. 
\section{Relevance}
\label{sec:relevance}

\textit{Relevance to the research community and society:}
Question Answering (QA) \cite{HoffnerWMULN17} is a crucial technology to provide intuitive end-users access to semantic data. Research on the development of QA applications is of interest to several scientific communities, including Semantic Web, Natural Language Processing, and Human Computer Interaction \cite{shekarpour2016question}. Semantic QA approaches automatically translate user queries posed in a natural language into structured queries, e.g., in the SPARQL query language. Whereas current research mostly focuses on entity-centric questions, events are still underrepresented in the semantic sources and the corresponding QA datasets.

\textit{Relevance for Question Answering applications:}
Event-centric semantic reference sources are still rare, with the first event-centric knowledge graphs such as EventKG evolving only recently. Often, event-centric information spread across entity-centric knowledge graphs is less annotated and more complex than the entity-centric information and is more difficult to retrieve. 
As a consequence, existing QA datasets such as LC-QuAD \cite{TrivediMDL17, lcquad2} and QALD are mainly entity-centric.
Specialized event-centric QA datasets are currently non-existing. The provision of such resources can bring a novel perspective in the Question Answering research and facilitate further development of QA approaches in the context of events.

\textit{Impact on the adoption of semantic technologies:}
Event-centric information is of crucial importance for researchers and practitioners in various domains, including journalists, media analysts, and Digital Humanities researchers. 
Current and historical events of global importance such as the COVID-19 outbreak, the Brexit, the Olympic Games and the US withdrawal from the nuclear arms treaty with Russia are the subject of current research in several fields including Digital Humanities and Web Science (see, e.g., \cite{GottschalkBRD18, Rogers:2013}). Event-centric repositories and the corresponding QA systems can help to answer relevant questions and support these studies.
We believe that the provision of intuitive access methods to the semantic reference sources can facilitate the broader adoption of semantic technologies by researchers and practitioners in these fields.

\section{Problem Statement}
\label{sec:problem}

Semantic Question Answering (QA) is a process of translating 
user questions expressed in a natural language into the 
corresponding semantic queries for a given knowledge graph. 

This work aims to create a Question Answering dataset to support the development and evaluation of QA approaches for event-centric questions. 

In the context of this work, events are real-world happenings of societal importance, 
such as the "2018 FIFA World Cup" and "2016 United States presidential election". 
Such events are typically found in encyclopedic sources like Wikipedia and the corresponding knowledge graphs such as EventKG \cite{Gottschalk:2018}, DBpedia \cite{dbpedia-swj} and Wikidata \cite{Vrandecic:2012}. Examples of relevant event types include military conflicts, sports tournaments, and political elections. In such datasets, events are typically modelled using semantic event models, as for example the Simple Event Model \cite{vanhage:2011} and the CIDOC CRM \cite{doerr2003cidoc}. 
Note that daily happenings represented as news headlines or unstructured textual descriptions, as in GDELT \cite{Leetaru:2013} and the Wikipedia Current Events Portal\footnote{\url{https://en.wikipedia.org/wiki/Portal:Current_events}} datasets are not in the scope of this work.

In the following, we define the relevant concepts and discuss requirements for an event-centric \textit{Event-QA} dataset. 

\begin{definition}\textbf{Knowledge Graph.}
\label{def:kg}
A knowledge graph $KG$ is a labelled multi-graph $KG=(V,R_v,R_l,L)$. 
$V=E_{v} \cup E_{n}$ is a set of nodes in $KG$. 
The set $E_v$ represents real-world events.
The set $E_n$ represents real-world entities.
$L$ is a set of literals. 
$R_{v}$ and $R_{l} $ are sets of edges. An edge in $R_v$ connects two nodes in $V$ and an edge in $R_l$ connects a node in $V$ with a literal in $L$.
\end{definition}

Following this definition, an event is represented as a node in $E_v$ in a knowledge graph. 
Literals represent specific properties of events and entities, e.g., the start time of the name.

\begin{definition}\textbf{Relation.}
\label{def:relation}
$Rel \subseteq V \times R_v \times V$ denotes the set of real-world relations between events and entities in a knowledge graph $KG$.
\end{definition}

Relation examples include sub-event relations (e.g., the "2018 FIFA World Cup Final" and the "2018 FIFA World Cup"), as well as relations connecting an event to its participants (e.g., the "French National football team" and the "2018 FIFA World Cup Final").

A \textit{query graph} is a subgraph of the knowledge graph. A query graph includes a subset of nodes and edges of the knowledge graph, 
as well as a set of variables representing such nodes and edges. 
As the focus of this work is on event-centric questions, 
at least one node in the query graph represents an event.

%
%

\begin{definition} \textbf{Query graph.}
\label{def:querygraph}
A query graph $q=(V',R_v', R_l', L',U)$ is a sub-graph of the knowledge graph $KG=(V,R_v,R_l,L)$, $V=E_{v} \cup E_{n}$,  
where $V' \subset V$, $R_v' \subset R_v$, $R_l' \subset R_l$, 
$L' \subset L$, and $U$ is a set of variables. 
Each variable $u \in U$ maps to a node of the knowledge graph $KG$.  
At least one node in the query graph represents an event: 
$\exists v'\in E_v:   v' \in  V'  \lor  \exists u' \in U: u' \mapsto v'$.
\end{definition}

A semantic query $q$ includes a query graph, a query type, and optionally a set of constraints.
The query type represents a projection operator such as SELECT and ASK 
or an aggregation operator such as COUNT. 
Constraints can, e.g., be used to restrict the period of 
interest.

\begin{definition} \textbf{Semantic query.}
\label{def:semanticquery}
A semantic query consists of: 
1) a query graph, 
2) a query type, 
3) an optional set of constraints.
\end{definition}

A semantic query can be expressed in the SPARQL query language.

In a QA dataset, each semantic query $q$ is aligned with one or more verbalizations, i.e., questions expressed in natural language(s).

In order to facilitate an adequate assessment of the performance of 
QA systems for event-centric queries, the QA dataset has to include QA tasks of sufficient difficulty. In particular, QA dataset should address the following issues: 
\begin{itemize}[topsep=0pt]
\item The complexity of semantic queries, i.e., the queries should include more than one triple pattern.
\item The diversity of the semantic queries, i.e., the semantic queries in the dataset should be dissimilar to each other. 
\item The diversity of the query verbalizations, i.e., the verbalizations of the queries should be dissimilar. 
\end{itemize}

\section{Event-QA Generation Approach} 
\label{sec:approach}

In this section, we present our approach to generating an event-centric QA dataset 
containing complex and diverse queries given a knowledge graph. 

\subsection{Semantic Query Generation Pipeline}
\label{sec:query_generation}

\begin{figure*}[ht]
    \centering
    \includegraphics[width=\textwidth]{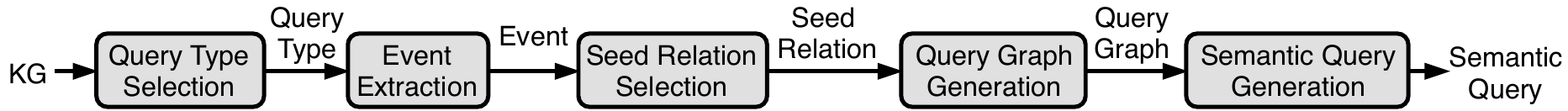}
    \caption{Overview of the \textit{Event-QA} pipeline. Given a knowledge graph as input, one execution of this pipeline leads to the generation of one semantic query.}
    \label{fig:pipeline}
\end{figure*}

We illustrate our semantic query generation pipeline in Figure \ref{fig:pipeline}.  

For each query to be generated, the \textit{Event-QA} pipeline includes the following steps: 

\begin{enumerate}
    \item \textbf{Query Type Selection:} A query type is selected randomly (i.e., ASK, SELECT or COUNT). 
   \item \textbf{Event Extraction:} A named event node from the knowledge graph is randomly selected together with all relations connected to that node. 
\item \textbf{Seed Relation Selection:} A relation is randomly selected from the list of all relations involving the 
previously selected event. 
We refer to such relations as seed relations. The seed relation includes at least one event, as ensured by step 2. 
    \item \textbf{Query Graph Generation:} A query graph is generated as a sub-graph of the knowledge graph augmented with variables. 
    \begin{enumerate}
        \item \textbf{Sub-Graph Generation:} To generate a sub-graph containing more than one relation, we conduct a random walk over the knowledge graph starting from the seed relation. 
        \item \textbf{Augmentation with Variables:} The sub-graph is complemented with variables to obtain a query graph. 
    \end{enumerate}
    \item \textbf{Semantic Query Generation:} 
    We augment the query graph with the query type and 
    optionally temporal constraints to build the semantic query.
    We translate the resulting semantic query into the SPARQL query language.
    \item \textbf{Query Verbalization:} For each SPARQL query resulting from the \textit{Event-QA} pipeline, the corresponding verbalization is created. 
\end{enumerate}

Figure \ref{fig:seed_relation} illustrates an example of seed relation. Based on this seed relation, we can create the query graph shown in Figure \ref{fig:query_graph} with a random walk based approach. Finally, we translate the query graph into a SPARQL query for a knowledge graph (Figure \ref{fig:sparql_query}).  

In order to generate a set of meaningful complex and diverse queries, we adopted a two-stage approach. First, we employed an initial version of the semantic query generation pipeline to automatically generate an initial sample of $100$ semantic queries. 
Using these queries, we conducted the first annotation stage, where we manually created verbalizations of the suggested semantic queries and marked the queries that did not appear meaningful. We analyzed the patterns in the annotations and used the collected observations to fine-tune the query generation pipeline. Finally, we used the pipeline to generate the final set of queries. Subsequently, we manually annotated this set to obtain the query verbalizations in the \textit{Event-QA} dataset.

In the following, we present the individual steps of the 
pipeline in more detail.

\subsection{Query Type Selection}
\label{sec:query-type-selection}
The pipeline starts with a random selection of the type for the semantic query to be generated. 
The query types included in the \textit{Event-QA} dataset are the SPARQL query forms ASK and SELECT\footnote{\url{https://www.w3.org/TR/sparql11-query}}, as well as the aggregation operator COUNT. 
ASK queries determine whether a query pattern has a solution.
SELECT returns variables and their bindings.
COUNT computes the number of results for a given expression.

\subsection{Event Extraction} 
\label{sec:evextraction}

To ensure that all queries in the dataset include at least one event, we start the query graph generation process by randomly picking one event from the event set $E_v$. This event and its associated relations build the input for the next step.

\subsection{Seed Relation Selection} 
\label{sec:seedselection}
\begin{figure}
  \centering
  \includegraphics[scale=0.55]{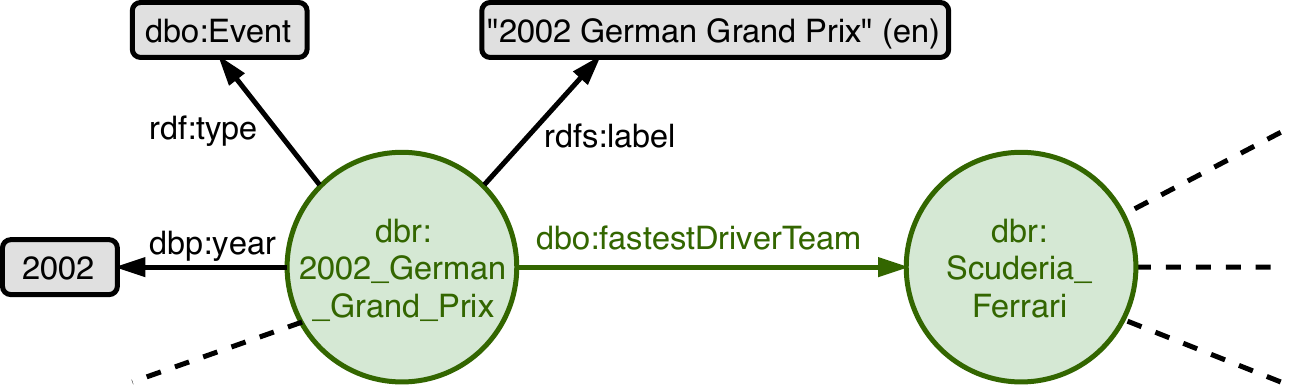}
  \caption{Seed relation example. Given the event labelled ``2002 German Grand Prix'' as the starting point, the relation $($\texttt{dbr:2002\_German\_Grand\_Prix}$,$ \texttt{dbo:fastestDriverTeam}$,$ \texttt{dbr:Scuderia\_Ferrari}$)$ (marked in green) was randomly selected from the knowledge graph as a seed relation.}
  \label{fig:seed_relation}
\end{figure}

We randomly choose a relation $(n_1, r, n_2) \in Rel$ that belongs to the set of relations referring to the event picked in the previous step. This relation takes the role of the \textit{seed relation} in the current execution of the \textit{Event-QA} pipeline.
As our goal is to generate complex event-centric queries, a seed relation needs to fulfill specific criteria: 
(i) the seed relation needs to include at least one event ($n_1 \in E_v \vee n_2 \in E_v$) which is guaranteed by step 2 in our pipeline, and (ii) at least one of the nodes included in the relation needs to be part of another relation in the knowledge graph.
Figure \ref{fig:seed_relation} provides an example of a seed relation $($\texttt{dbr:2002\_German\_Grand\_Prix}$,$ \texttt{dbo:fastestDriverTeam}$,$ \texttt{dbr:Scuderia\_Ferrari}$)$\footnote{\texttt{dbr} is the prefix of the DBpedia resource identifier: http://dbpedia.org/resource/.}.

\subsection{Query Graph Generation} 
\label{sec:query-graph-generation}

Based on the seed relation extracted in the previous step, we create a query graph. This query graph is a sub-graph of the knowledge graph that contains the seed relation and variables. Thus, two steps are required to create the query graph: (i) sub-graph generation, and (ii) augmentation with variables.

\subsubsection{Sub-Graph Generation}

We obtain a sub-graph of the knowledge graph through a random walk procedure.
The seed relation constitutes the initial sub-graph.
Through a random walk, we incrementally extend the sub-graph by adding new relations to the nodes already included in the sub-graph in a way that the sub-graph remains connected. 
In each step, we randomly select a node of the sub-graph. Then, we randomly select an edge from the knowledge graph connected to that node. If this edge is not part of the sub-graph yet, we add the corresponding relation to the sub-graph.

\begin{figure}
  \centering
  \includegraphics[scale=0.55]{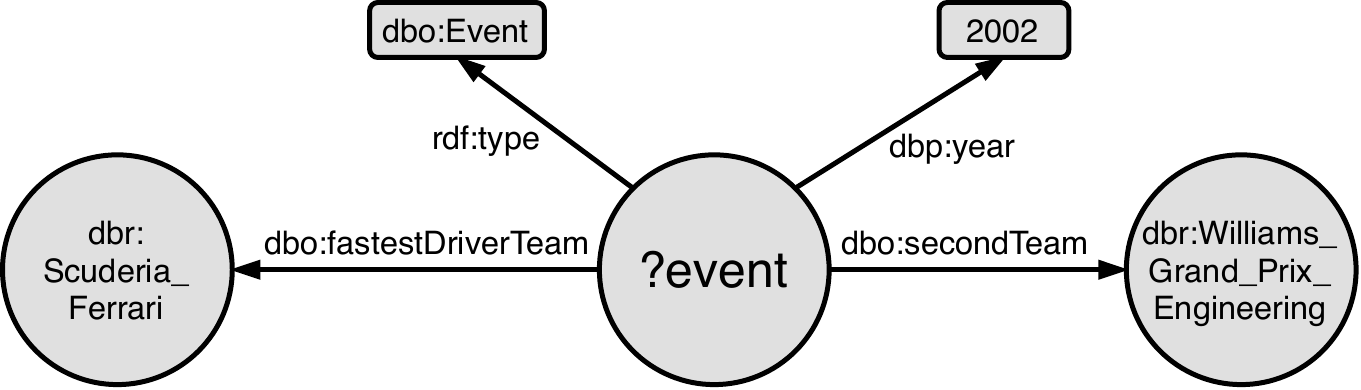}
  \caption{Query graph example. Following the random walk, another edge was added to the initial sub-graph shown in Figure \ref{fig:seed_relation}. 
  The event node was replaced with a variable.}
  \label{fig:query_graph}
\end{figure}

Figure \ref{fig:query_graph} is an example of a query graph from the seed relation in Figure \ref{fig:seed_relation}. Another relation was added $($\texttt{?event}$,$ \texttt{dbo:\allowbreak secondTeam}$,$ \texttt{dbr:\allowbreak Williams\allowbreak \_Grand\allowbreak \_Prix\allowbreak \_Engineering}$)$. The variable \texttt{?event} is mapped to \texttt{dbr:\allowbreak 2002\allowbreak \_German\allowbreak \_Grand\allowbreak \_Prix}. Additionally, the event labelled ``2002 German Grand Prix'' was replaced with a variable 
(we describe the methods for selecting such variables in more detail later in Section \ref{sec:aug-variables}). The resulting query graph can be represented in natural language as: 
\textit{In which competition in 2002 did Ferrari appear as the fastest driver team and Williams as the second team?}

The random walk continues until the algorithm meets the termination condition. In particular, we apply a threshold to restrict the maximal number of relations of the query. The value of this threshold is decided based on the manual annotations in the fine-tuning stage. Here we observed that the majority of the queries that included three or more relations were difficult to interpret. 

We illustrate the issue of high complexity at the example from Figure \ref{fig:query_graph}. 
With the threshold value above two, a third relation, e.g., \texttt{?event dbo:secondDriver dbr:Juan\_Pablo\_Montoya},  is added to the query graph. 
The corresponding English verbalization of this query could be as follows: 
\textit{In which competition in 2002, where Juan Pablo Montoya was the second driver, did Ferrari appear as the fastest driver team and Williams as the second team?}
The participants annotated this query as too complicated during the first annotation round.

Therefore, in the current version of the \textit{Event-QA} dataset, we restrict the threshold value to two\footnote{Note that this limit does not include temporal constraints, meaning that a query can contain two relations and additional temporal constraints.}. 
In principle, the proposed approach is flexible concerning the threshold value such that semantic queries with 
more than two relations can be generated in cases where such configurations are meaningful, e.g., in domain-specific knowledge graphs.  

\subsubsection{Augmentation with Variables}
\label{sec:aug-variables}

In this step, we systematically select the variable $u \in U$ to build the query graph, given the sub-graph resulting from the previous step. 
A variable can replace either a literal or a node in the query graph. 
We assign variables randomly. 
Figure \ref{fig:query_graph} provides an example of a resulting query graph. 

Whereas we experimented with multiple variables per query graph and an entirely random assignment procedure, our observations of the annotations during the pipeline tuning resulted in constraints for the variable selection. These constraints increase the likelihood that the resulting queries are meaningful.  

In particular, in the final pipeline configuration, we: 
\begin{itemize}
    \item add at most one variable to the query graph; 
    \item do not include variables to the ASK queries; 
    \item avoid variables as leaf nodes of the query graph; 
    \item avoid variables representing information redundant in the query graph. For example, DBpedia sometimes includes time information as event name part, e.g., "\textit{2002 German Grand Prix}". In these cases, we should avoid variables representing event time.
    \item avoid variables representing literals containing time-related information in COUNT queries.
\end{itemize}

\subsubsection{Augmentation with Temporal Constraints} 
\label{sec:temp-constraints}

When possible, we include temporal constraints. Such temporal information can, for example, denote the validity time of a relation or an existence time of an entity. Based on this information, we define a time interval of interest. Then, we randomly select one of the following constraints and add it to the query: (i) within the time interval, (ii) after the time interval, or (iii) before the time interval.

\subsection{Semantic Query Generation} 

Together, query type, query graph, and optional temporal constraints constitute a semantic query. 
We automatically translate this semantic query into a SPARQL query
for the specific knowledge graph.

Figure \ref{fig:sparql_query} provides an example of a SPARQL query, 
given the query graph in Figure \ref{fig:query_graph} with a temporal constraint.

\begin{figure}

\begin{lstlisting}[basicstyle=\ttfamily,language=sql,deletekeywords={count,YEAR,INTEGER},keywordstyle=\bfseries,frame=single,sensitive=t,morekeywords={FILTER,COUNT}]
SELECT (COUNT(DISTINCT(?event) AS ?count)) WHERE {
  ?event rdf:type dbo:Event .

  ?event dbo:fastestDriverTeam dbr:Scuderia_Ferrari .
  ?event dbo:secondTeam dbr:Williams_Grand_Prix_Engineering .

  ?event dbp:year ?year .
  FILTER ( ?year > "2001"^^xsd:integer)
}
\end{lstlisting}
\caption{SPARQL query example for the query graph illustrated in Figure \ref{fig:query_graph} for the DBpedia knowledge graph.} 
\label{fig:sparql_query}
\end{figure}

\subsection{Pipeline Fine-tuning} 
\label{sec:pipeline_tuning}
The pipeline fine-tuning and the creation of query verbalizations (i.e., natural language representations of semantic queries) in the \textit{Event-QA} dataset are conducted manually. The authors of this work and three post-graduates with expertise in SPARQL and RDF participated in the verbalization. To facilitate these tasks, we implemented a Web interface that displayed the SPARQL queries generated by the pipeline shown in Figure \ref{fig:pipeline}.

In order to collect input for the pipeline fine-tuning, we used the interface to observe any systematic patterns that make the queries difficult to understand or difficult to translate into meaningful natural language questions. Our interface provided the following instructions for annotation of a given SPARQL query:
\begin{enumerate}
    \item Read the SPARQL query and think of the question it represents.
    \begin{itemize}
        \item If you do not understand the query, select the "I do not understand the query" option. Leave a comment on what makes it difficult to understand the query. Click "continue". The next query will be shown.
    \end{itemize}
    \item Do you think that a human user would ask the question represented by this query?
       \begin{itemize}
        \item If you think that is a question a user would not ask, please select the option "A user would not ask this question". Leave a comment to explain why. Click "continue". The next query will be shown.
    \end{itemize}
\end{enumerate}

From this annotation process on 100 queries, we have gained insights into the query complexity and the allocation of variables.

\textbf{Query Complexity:} As stated in Section \ref{sec:query-graph-generation}, we observed that queries which included more than two relations were less meaningful and more difficult to understand for the participants. 
Although in general, more complex queries can be meaningful, 
we did not observe such examples in our dataset. 
As a result, we restricted the complexity of the queries in the dataset to a maximum of two relations.

\textbf{Allocation of Variables:} An appropriate allocation of variables is one of the most critical issues to automatically generate meaningful queries. 
In particular, queries containing multiple variables
are often difficult to understand or do not have any meaningful interpretation. The same observation applies to the ASK queries that contain variables. 
Finally, queries that include multiple relations and 
contain variables at the leaf nodes of the query graph 
do not result in a meaningful natural language interpretation. 
Instead, the nodes that connect several relations in the query graph are more suitable for variables allocation. 
For the creation of the final set of queries, we introduced the corresponding rules in the variable augmentation step of the semantic query generation pipeline described in Section \ref{sec:aug-variables}.

\subsection{Query Verbalization} 
\label{sec:query_verbalisations}

The SPARQL queries generated using the fine-tuned semantic query generation pipeline are annotated with English, Portuguese and German questions. 
For the English questions, each SPARQL query verbalization was manually confirmed as in the first annotation step described in Section \ref{sec:pipeline_tuning}. 
For the annotations, we formulated the following instructions:
\begin{itemize}[topsep=0pt]
\item Formulate the question in a way that sounds natural.
\item If possible, vary the language expressions you use for different queries.
\end{itemize}
Finally, native Portuguese and German speakers among the authors provided high-quality translations of the English queries in the corresponding language. 

In principle, the translation of NL expressions from SPARQL queries and the translation of NL expressions across languages can be automated using existing approaches such as SPARQL2NL \cite{NgomoBULG13a} and machine translation. However, we observed two main problems. First, NL-expressions automatically generated by the state-of-the-art tools do not result in intuitive sentences for complex event-centric SPARQL queries in our dataset. Second, we tried to use such automatically generated NL-expressions as an initial suggestion for manual post-editing. We observed that these suggestions do not help to speed up the manual translation process, as the annotators have to understand a complicated NL-expression in addition to the original SPARQL query. Therefore, to ensure the NL-quality and the manual annotation process's efficiency, we stick to the manual query verbalization. In future work, we will investigate how automatic translation methods can support the \textit{Event-QA} pipeline and to which extent further development of the automatic translation methods is required.

\section{Application to EventKG}
\label{sec:implementation}

In this section, we present the application of 
the QA dataset generation method described in Section \ref{sec:approach}
to the EventKG knowledge graph. 

\subsection{EventKG Knowledge Graph}
\label{sec:eventkg}

The knowledge graph adopted for the creation of queries in \textit{Event-QA} is EventKG \cite{Gottschalk:2018} -- a multilingual large-scale temporal knowledge graph. 
EventKG V2.0, released in 03/2019, builds a basis for the \textit{Event-QA} generation and contains over 970k contemporary and historical events and over 2.8 million temporal relations extracted from DBpedia, Wikidata, and YAGO knowledge graphs as well as several semi-structured sources.

EventKG can be directly expressed as a knowledge graph according to Definition \ref{def:kg}, using EventKG's ontology that is based on the Simple Event Model (\texttt{sem}) \cite{vanhage:2011}. The set of events $E_v$ consists of all EventKG resources typed as \texttt{sem:Event}, the set of entities $E_n$ consists of all EventKG resources typed as an instance of \texttt{sem:Core}, but not as \texttt{sem:Event}. The set of relations $R_v$ corresponds to EventKG's \texttt{eventKG-s:Relation} instances\footnote{In EventKG, most relations are modelled as resources that point to the relation's subject (via \texttt{rdf:subject}), object (\texttt{rdf:object}) and property (\texttt{sem:roleType}).} and any other relations expressed using predefined properties such as \texttt{dbo:nextEvent} and \texttt{sem:hasPlace}. In EventKG, begin and end times of events and existence times of entities are represented by \texttt{sem:hasBeginTimeStamp} and \\
\texttt{sem:hasEndTimeStamp} properties. These relations constitute the set of relations $R_l$.

\subsection{Applying the \textit{Event-QA} Query Generation Algorithm to EventKG}
\label{sec:event-kg-application-example}

\begin{figure} [h]
    \centering
    \includegraphics[scale=0.5]{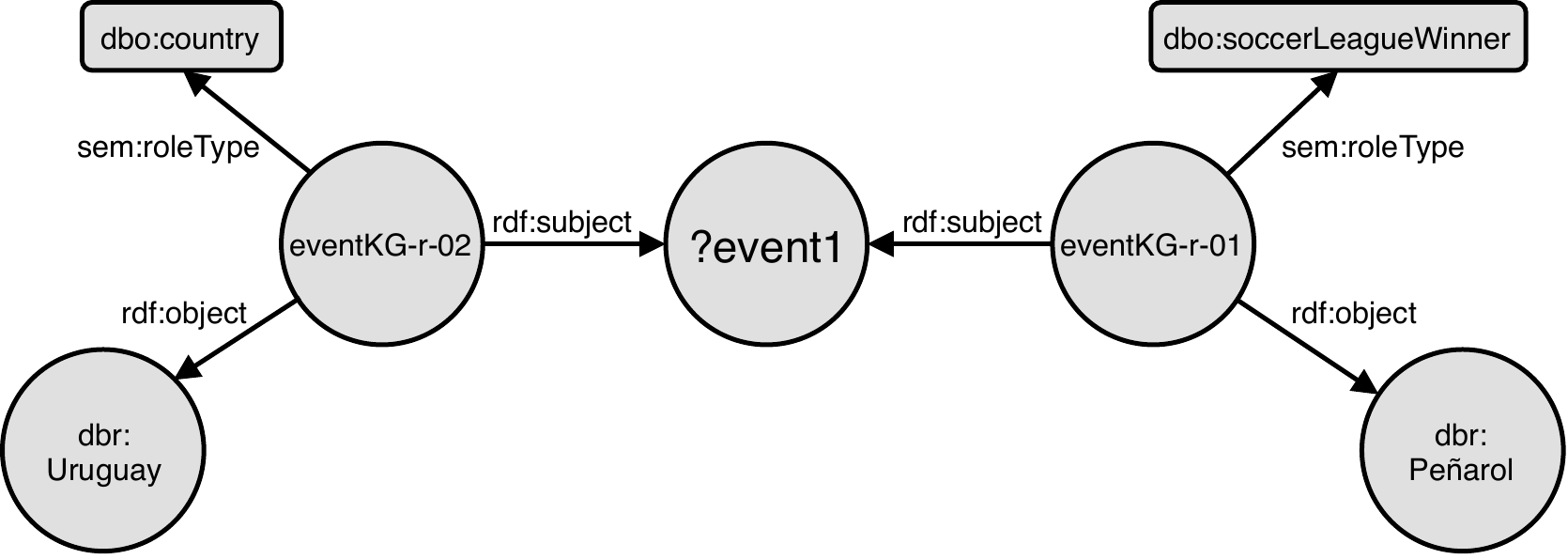}
    \caption{An example of a query graph created when applying the \textit{Event-QA} algorithm to EventKG. For readability, we denote the nodes using their DBpedia resource identifiers (\texttt{dbr}).}
    \label{fig:subgraph_eventkg}
\end{figure}

Figure \ref{fig:subgraph_eventkg} illustrates an example of how the \textit{Event-QA} query graph generation approach is applied to the EventKG knowledge graph.
\texttt{eventKG-r-01} takes the role of the seed relation in this example.
This relation connects an event node (e.g., \texttt{dbr:1973\_Uruguayan\_Primera\_División}) with the entity \texttt{dbr:Peñarol}. 

In this example, the first application of the random walk leads to the selection of the event node, connected to the relation node \texttt{eventKG-r-02}. In the next iteration, the relation node \texttt{eventKG-r-02} is added to the sub-graph, which connects the sub-graph to the entity \texttt{dbr:Uruguay}. 

To obtain the query graph, a randomly selected node of the sub-graph is replaced with a variable. 
In the example illustrated, the event node plays the role of a variable which results in the SPARQL query shown in Figure \ref{fig:sparql_eventkg_query}. This query aims to select soccer leagues (\texttt{dbo:\-soccer\-League\-Winner}) that happened (\texttt{dbo:country}) in Uruguay (\texttt{dbr:Uruguay}) and that were won (\texttt{dbo:\-soccer\-League\-Winner}) by Peñarol (\texttt{dbr:Peñarol}), i.e., the query can be verbalised as: \textit{"In which soccer leagues did Peñarol win in Uruguay?".}

\begin{figure}
\begin{lstlisting}[basicstyle=\ttfamily,deletekeywords={year},frame=single,inputencoding=utf8,extendedchars=true,
    literate={ñ}{{\~n}}1]
SELECT DISTINCT ?event WHERE {
  ?relation1 rdf:object ?entity1 .
  ?relation1 rdf:subject ?event .
  ?relation1 sem:roleType dbo:country .

  ?relation2 rdf:object ?entity2 .
  ?relation2 rdf:subject ?event .
  ?relation2 sem:roleType dbo:soccerLeagueWinner .

  ?entity1 owl:sameAs dbr:Uruguay .
  ?entity2 owl:sameAs dbr:Peñarol .
}
\end{lstlisting}
\caption{Translation of the query graph in Figure \ref{fig:subgraph_eventkg} into a SPARQL query for EventKG.}
\label{fig:sparql_eventkg_query}
\end{figure}

\subsection{Translation of EventKG Queries to DBpedia}
\label{sec:dbpedia-application}

As \textit{Event-QA} is a dataset with event-centric queries, it is expected to perform best on an event-centric knowledge graph such as EventKG. However, to facilitate an easier evaluation of existing QA systems using the \textit{Event-QA} dataset, we also provide a translation of the SPARQL queries to the English DBpedia knowledge graph due to its popularity in the existing QA systems, where possible. This translation is performed by transforming the \texttt{eventKG-s:Relation} instances to the DBpedia triples using the \texttt{sem:roleType} values. Whenever possible, start and end dates from EventKG are mapped to the temporal DBpedia predicates (e.g., \texttt{dbp:year}, \texttt{dbo:date} and \texttt{dbo:startDate}). 

$307$ out of $1000$ SPARQL queries that target EventKG in \textit{Event-QA} can be translated to the English DBpedia. 
This number can be explained by the underlying structural and content differences between these two knowledge graphs. DBpedia is not specialized in events and covers much less event-centric information compared to EventKG. The differences are as follows:  
(i) Semantic queries generated using EventKG contain relations that originate from a variety of sources (e.g., Wikidata and non-English DBpedia versions); these relations are not always present in the English DBpedia. (ii) Event instances are underrepresented in the English DBpedia, compared to EventKG. (iii) DBpedia does not contain unified dedicated temporal properties so that the automatic mapping of temporal properties for the individual entities and events often fails.

\section{Evaluation and Dataset Characteristics}
\label{sec:evaluation}

The quality of queries and the correctness of the translation between SPARQL and natural language is essential towards the quality of a Question Answering dataset. In this section, we describe our methods to ensure dataset quality and provide dataset statistics and example queries. Additionally, we provide insights into the diversity and complexity of the \textit{Event-QA} dataset.

\subsection{Dataset Quality}
In order to assess the correctness of the translation between the SPARQL queries and the corresponding natural language expressions, we manually verified all the queries included in the dataset. We corrected verbalizations in case of any issues. 
This way, we ensure that verbalizations correctly represent the questions expressed by the SPARQL queries.
 
\subsection{Dataset Statistics and Examples}

\textit{Event-QA} consists of $1000$ verified semantic queries. Figure \ref{fig:example_queries} lists three example query verbalizations that are part of the dataset.

\begin{figure}[th]
\footnotesize
\begin{center}
\begin{tabular}{l}
\hline
\begin{tabular}[c]{@{}l@{}}
When did the Excitante music festival finish in Argentina? \\
$\hookrightarrow$ PT: Quando o festival de música Excitante terminou na Argentina?\\ $\hookrightarrow$ DE: Wann ging das argentinische Musical Excitante zu Ende?
\end{tabular} \\ \hline
\begin{tabular}[c]{@{}l@{}}Give me a list of football games won by Dynamo Kyiv.
\end{tabular} \\ \hline
How many events chaired by Derek Shaw were part of the EFL championships? \\ \hline
\end{tabular}\caption{Example natural language questions in \textit{Event-QA}. For the first example, we also show the Portuguese and the German translation.}
\label{fig:example_queries}
\end{center}
\end{figure}

In total, $1005$ different events, $1655$ different entities and $309$ different relations occur within the $1000$ SPARQL queries.
The most frequent relations are  \texttt{dbo:commander}, \texttt{dbo:award}, \texttt{dbo:city}, \texttt{dbo:battle}, \texttt{dbo:birth\-Place}, and 
\voc{dbo}{sport}.

\subsection{Complexity and Diversity}
\label{sec:metrics}

To assess to which extent the dataset generated in this work satisfies the conditions specified in Section \ref{sec:problem}, we define metrics to assess the complexity and the diversity of semantic queries as well as the diversity of their verbalizations. For comparison, we compute the values of such metrics for \textit{Event-QA}, as well as for the QA datasets LC-QuAD \cite{TrivediMDL17}, Saquete et al. \cite{SaqueteGMML09} and TempQuestions \cite{TempQuestions}. The results are shown in Table \ref{tab:results} and are explained here in more detail. Overall, the datasets behave similarly concerning the complexity and diversity metrics, whereas only \textit{Event-QA} focuses on events.

\begin{table*}[ht]
\centering
\caption{Complexity, query diversity, and verbalization diversity (in English, Portuguese and German) of \textit{Event-QA} in comparison to other QA datasets. Complexity and query diversity are only reported if SPARQL queries are available.}
\label{tab:results}
\begin{tabu}{c|[1.5pt]c|c|c}
\textbf{Dataset}  & \textbf{Query Complexity} & \multicolumn{1}{c|}{\begin{tabular}[c]{@{}c@{}}\textbf{Query Diversity}\end{tabular}} & \textbf{Verbalization Diversity} \\ \tabucline[1.5pt]{-}
LC-QuAD  & 2.0                             & 0.95                                                 & 0.87                                                \\ \hline
\textit{Event-QA} & 2.0                    & \textbf{0.98}                                        & 0.82 (EN), 0.86 (PT), 0.87 (DE)                     \\ \hline
Saquete et al. \cite{SaqueteGMML09} & -                    & -                                        & 0.84                     \\ \hline
TempQuestions \cite{TempQuestions} & -                    & -                                        & \textbf{0.89}                     \\ 

\end{tabu}
\end{table*}

\subsubsection{Query Complexity}

The complexity of a semantic query grows with an increasing number of relations. For simplicity, we assume linear growth. Based on this assumption, we measure the complexity of a semantic query as the number of relations included in the query graph. The overall complexity of the dataset is computed as an average complexity of the queries it contains. 
Table \ref{tab:results} shows that \textit{Event-QA} queries have $2$ relations on average and thus are as complex as the LC-QuAD queries.
%

\subsubsection{Query Diversity}

The dataset $\beta$ is semantically diverse if its semantic queries are dissimilar. 
Similar queries share events, entities, and relations. Therefore, we assess the similarity of semantic queries using the Jaccard coefficient applied to the set of nodes and edges contained in the 
corresponding query graphs. To assess the diversity of the queries in the dataset as a whole, 
we compute the diversity as an average dissimilarity value across all query pairs:

\begin{equation}
    diversity(\beta) = 1- \frac{ \sum_{(q_i, q_j) \in \beta} similarity (q_i, q_j) }{\binom{|\beta|}{2}},
\end{equation}

\noindent where $(q_i, q_j)$ is a query pair in the dataset with $q_i \neq q_j$  and $|\beta|$ is the number of queries in the dataset.

Semantic queries in \textit{Event-QA} are slightly more diverse than those from LC-QuAD, as per Table \ref{tab:results}. 
Differently from LC-QuAD, our approach does not depend on templates to build semantic queries. Instead, query graphs in \textit{Event-QA} 
are build by randomly taken decisions (i.e., random seed selection and random walk). 

\subsubsection{Verbalization Diversity} 

The dataset $\beta$ is textually diverse if the verbalizations $q_{NL}$ of its queries are dissimilar. 
Similar verbalizations share terms. Therefore, we assess the similarity of the verbalizations using the cosine similarity. 
To assess the diversity of query verbalizations in the dataset as a whole, 
we compute the diversity as an average dissimilarity value across all pairs of verbalizations:

\begin{equation}
    diversity(\beta_{NL}) = 1- \frac{ \sum_{(q_{NL_i}, q_{NL_j}) \in \beta} cos(q_{NL_i}, q_{NL_j}) }{\binom{|\beta|}{2}},
\end{equation}

\noindent where $(q_{NL_i}, q_{NL_j})$ is a verbalization 
pair in the dataset and $|\beta|$ is the number of queries in the dataset.

Table \ref{tab:results} reveals that the verbalizations of TempQuestions are most diverse, which was expected, given that this dataset combines three different source datasets. \textit{Event-QA}'s English and Portuguese query verbalizations are more diverse than in Saquete et al. and only slightly less diverse than in LC-QuAD. In summary, our results show that concerning the complexity and diversity, \textit{Event-QA} performs similar to other state-of-the-art QA datasets.

\section{Availability \& Sustainability}
\label{sec:availability}

The \textit{Event-QA} homepage\footnote{\url{http://eventcqa.l3s.uni-hannover.de/}} provides a description of the dataset and how to use and cite the resource. 
In addition, we provide permanent access to the data on 
Zenodo\footnote{\url{https://doi.org/10.5281/zenodo.3568387}} released under CCBY 4.0\footnote{\url{https://creativecommons.org/licenses/by/4.0/}}. The DOI of the dataset is: \textbf{10.5281/zenodo.3568387}. The data we provide includes:

\begin{itemize}
    \item \textbf{Event-QA}: Event-Centric Question Answering dataset in the QALD JSON format \cite{UsbeckGN018} for EventKG. For each query we provide a SPARQL expression for EventKG (and where possible for DBpedia), verbalizations in English, Portuguese and German, and gold-standard answers.
    \item \textbf{Predicates}: List of DBpedia predicates used in the dataset.
    \item \textbf{VoID description}: Machine readable description of the dataset in RDF.
    \item List of DBpedia \textbf{events} and \textbf{entities} covered by the dataset.
\end{itemize}

The dataset generation pipeline is available as open source software on GitHub\footnote{\url{https://github.com/tarcisiosouza/Event-QA}} under the MIT License\footnote{\url{https://opensource.org/licenses/MIT}}.

The \textit{Event-QA} dataset was initially developed using EventKG V2.0, released in 03/2019. 
\textit{Event-QA} is compatible with the new version of the knowledge graph, EventKG V3.0 released in 03/2020.
Regarding sustainability, we plan to provide updates to ensure compatibility with future versions of EventKG. 
We plan to integrate \textit{Event-QA} into the FAIR benchmarking platform GERBIL QA\footnote{\url{http://gerbil-qa.aksw.org/gerbil/}}, which powered past QALD challenges.

\balance 

\section{Related Work}
\label{sec:related}
Semantic Question Answering over knowledge graphs, with recent approaches including IQA \cite{ZAFAR2020} and WDAqua \cite{DiefenbachBSM20}, is a difficult problem \cite{HoffnerWMULN17}. In order to evaluate semantic Question Answering approaches, a number of datasets that contain natural language questions and the corresponding SPARQL queries has been recently developed. 
Existing QA datasets greatly vary in size, content (question/answer pairs vs. semantic queries), complexity (i.e., number of triple patterns in the semantic queries), coverage of natural languages, quality of natural language, and target knowledge bases. 
Even more importantly, existing QA datasets for complex questions such as LcQuAD \cite{TrivediMDL17, lcquad2} do not sufficiently address questions regarding events. This shortcoming can be partially attributed to the fact that most of the popular KGs are entity-centric, whereas event-centric KGs such as EventKG evolved only recently. Although a few QA datasets (TempQuestions \cite{TempQuestions} and Saquete et al. \cite{SaqueteGMML09}) focus on temporal expressions, they do not sufficiently cover event-centric questions. 
Another important shortcoming of many existing datasets is the deficiency in natural language expression quality. In particular, we have observed that in cases where such expressions are generated through automated processes or crowd-sourcing with non-expert users (e.g., LCQuAD 2.0 \cite{lcquad2}, ComplexWebQuestions \cite{complexwebquestions}).

In contrast, Event-QA provides complex event-centric questions in three languages while adopting an automatic question-generation process and manual verbalization by expert users to ensure quality. With its focus on event-centric questions and manual verification of multilingual natural language expressions, Event-QA takes a unique position among the state-of-the-art QA datasets.

\section{Conclusion}
\label{sec:conclusion}

In this paper we presented \textit{Event-QA} -- an event-centric dataset for semantic Question Answering.
This dataset is generated through a random walk-based approach applied to the EventKG knowledge graph to ensure the diversity of the resulting queries. 
The translation of the resulting semantic queries into natural language is performed and verified manually to ensure the verbalization quality.  The resulting dataset contains 1000 complex event-centric questions that can be used for training and evaluation of semantic Question Answering systems. 
The query verbalizations in the \textit{Event-QA} are available in English, Portuguese, and German.

\graphicspath{ {./images/} }

\section*{Acknowledgments}
This work was partially funded by the European Research Council under
Cleopatra (H2020-MSCA-ITN-2018 812997). Tarcísio Souza Costa is sponsored by a scholarship from CNPq, a Brazilian government institution for scientific development.

\bibliographystyle{ACM-Reference-Format}
\bibliography{dl}


\begin{thebibliography}{22}


\ifx \showCODEN    \undefined \def \showCODEN     #1{\unskip}     \fi
\ifx \showDOI      \undefined \def \showDOI       #1{#1}\fi
\ifx \showISBNx    \undefined \def \showISBNx     #1{\unskip}     \fi
\ifx \showISBNxiii \undefined \def \showISBNxiii  #1{\unskip}     \fi
\ifx \showISSN     \undefined \def \showISSN      #1{\unskip}     \fi
\ifx \showLCCN     \undefined \def \showLCCN      #1{\unskip}     \fi
\ifx \shownote     \undefined \def \shownote      #1{#1}          \fi
\ifx \showarticletitle \undefined \def \showarticletitle #1{#1}   \fi
\ifx \showURL      \undefined \def \showURL       {\relax}        \fi
\providecommand\bibfield[2]{#2}
\providecommand\bibinfo[2]{#2}
\providecommand\natexlab[1]{#1}
\providecommand\showeprint[2][]{arXiv:#2}

\bibitem[\protect\citeauthoryear{Diefenbach, Both, Singh, and Maret}{Diefenbach
  et~al\mbox{.}}{2020}]%
        {DiefenbachBSM20}
\bibfield{author}{\bibinfo{person}{Dennis Diefenbach}, \bibinfo{person}{Andreas
  Both}, \bibinfo{person}{Kamal Singh}, {and} \bibinfo{person}{Pierre Maret}.}
  \bibinfo{year}{2020}\natexlab{}.
\newblock \showarticletitle{{Towards a Question Answering System over the
  Semantic Web}}.
\newblock \bibinfo{journal}{\emph{Semantic Web}} \bibinfo{volume}{11},
  \bibinfo{number}{3} (\bibinfo{year}{2020}), \bibinfo{pages}{421--439}.
\newblock


\bibitem[\protect\citeauthoryear{Doerr}{Doerr}{2003}]%
        {doerr2003cidoc}
\bibfield{author}{\bibinfo{person}{Martin Doerr}.}
  \bibinfo{year}{2003}\natexlab{}.
\newblock \showarticletitle{{The CIDOC Conceptual Reference Module: an
  Ontological Approach to Semantic Interoperability of Metadata}}.
\newblock \bibinfo{journal}{\emph{AI magazine}} \bibinfo{volume}{24},
  \bibinfo{number}{3} (\bibinfo{year}{2003}), \bibinfo{pages}{75--75}.
\newblock


\bibitem[\protect\citeauthoryear{Dubey, Banerjee, Abdelkawi, and Lehmann}{Dubey
  et~al\mbox{.}}{2019}]%
        {lcquad2}
\bibfield{author}{\bibinfo{person}{Mohnish Dubey}, \bibinfo{person}{Debayan
  Banerjee}, \bibinfo{person}{Abdelrahman Abdelkawi}, {and}
  \bibinfo{person}{Jens Lehmann}.} \bibinfo{year}{2019}\natexlab{}.
\newblock \showarticletitle{{LC-QuAD 2.0: A Large Dataset for Complex Question
  Answering over Wikidata and DBpedia}}. In
  \bibinfo{booktitle}{\emph{Proceedings of the International Semantic Web
  Conference}}. \bibinfo{pages}{69--78}.
\newblock


\bibitem[\protect\citeauthoryear{Gottschalk, Bernacchi, Rogers, and
  Demidova}{Gottschalk et~al\mbox{.}}{2018}]%
        {GottschalkBRD18}
\bibfield{author}{\bibinfo{person}{Simon Gottschalk}, \bibinfo{person}{Viola
  Bernacchi}, \bibinfo{person}{Richard Rogers}, {and} \bibinfo{person}{Elena
  Demidova}.} \bibinfo{year}{2018}\natexlab{}.
\newblock \showarticletitle{{Towards Better Understanding Researcher Strategies
  in Cross-Lingual Event Analytics}}. In \bibinfo{booktitle}{\emph{Proceedings
  of the International Conference on Theory and Practice of Digital
  Libraries}}. \bibinfo{pages}{139--151}.
\newblock


\bibitem[\protect\citeauthoryear{Gottschalk and Demidova}{Gottschalk and
  Demidova}{2018}]%
        {Gottschalk:2018}
\bibfield{author}{\bibinfo{person}{Simon Gottschalk} {and}
  \bibinfo{person}{Elena Demidova}.} \bibinfo{year}{2018}\natexlab{}.
\newblock \showarticletitle{{{EventKG}: A Multilingual Event-Centric Temporal
  Knowledge Graph}}. In \bibinfo{booktitle}{\emph{Proceedings of the Extended
  Semantic Web Conference}}. Springer, \bibinfo{pages}{272--287}.
\newblock


\bibitem[\protect\citeauthoryear{Gottschalk and Demidova}{Gottschalk and
  Demidova}{2019}]%
        {GottschalkD19}
\bibfield{author}{\bibinfo{person}{Simon Gottschalk} {and}
  \bibinfo{person}{Elena Demidova}.} \bibinfo{year}{2019}\natexlab{}.
\newblock \showarticletitle{{EventKG - the Hub of Event Knowledge on the Web -
  and Biographical Timeline Generation}}.
\newblock \bibinfo{journal}{\emph{Semantic Web}} \bibinfo{volume}{10},
  \bibinfo{number}{6} (\bibinfo{year}{2019}), \bibinfo{pages}{1039--1070}.
\newblock


\bibitem[\protect\citeauthoryear{H{\"o}ffner, Walter, Marx, Usbeck, Lehmann,
  and Ngonga~Ngomo}{H{\"o}ffner et~al\mbox{.}}{2017}]%
        {HoffnerWMULN17}
\bibfield{author}{\bibinfo{person}{Konrad H{\"o}ffner},
  \bibinfo{person}{Sebastian Walter}, \bibinfo{person}{Edgard Marx},
  \bibinfo{person}{Ricardo Usbeck}, \bibinfo{person}{Jens Lehmann}, {and}
  \bibinfo{person}{Axel-Cyrille Ngonga~Ngomo}.}
  \bibinfo{year}{2017}\natexlab{}.
\newblock \showarticletitle{{Survey on Challenges of Question Answering in the
  Semantic Web}}.
\newblock \bibinfo{journal}{\emph{Semantic Web}} \bibinfo{volume}{8},
  \bibinfo{number}{6} (\bibinfo{year}{2017}), \bibinfo{pages}{895--920}.
\newblock


\bibitem[\protect\citeauthoryear{Jia, Abujabal, Saha~Roy, Str\"{o}tgen, and
  Weikum}{Jia et~al\mbox{.}}{2018}]%
        {TempQuestions}
\bibfield{author}{\bibinfo{person}{Zhen Jia}, \bibinfo{person}{Abdalghani
  Abujabal}, \bibinfo{person}{Rishiraj Saha~Roy}, \bibinfo{person}{Jannik
  Str\"{o}tgen}, {and} \bibinfo{person}{Gerhard Weikum}.}
  \bibinfo{year}{2018}\natexlab{}.
\newblock \showarticletitle{{TempQuestions: A Benchmark for Temporal Question
  Answering}}. In \bibinfo{booktitle}{\emph{Proceedings of the Web
  Conference}}. \bibinfo{pages}{1057--1062}.
\newblock


\bibitem[\protect\citeauthoryear{Leetaru and Schrodt}{Leetaru and
  Schrodt}{2013}]%
        {Leetaru:2013}
\bibfield{author}{\bibinfo{person}{Kalev Leetaru} {and}
  \bibinfo{person}{Philip~A Schrodt}.} \bibinfo{year}{2013}\natexlab{}.
\newblock \showarticletitle{{{GDELT}: Global Data on Events, Location, and
  Tone, 1979-2012}}. In \bibinfo{booktitle}{\emph{ISA annual convention}},
  Vol.~\bibinfo{volume}{2}. Citeseer, \bibinfo{pages}{1--49}.
\newblock


\bibitem[\protect\citeauthoryear{Lehmann, Isele, Jakob, Jentzsch, Kontokostas,
  Mendes, Hellmann, Morsey, van Kleef, Auer, and Bizer}{Lehmann
  et~al\mbox{.}}{2014}]%
        {dbpedia-swj}
\bibfield{author}{\bibinfo{person}{Jens Lehmann}, \bibinfo{person}{Robert
  Isele}, \bibinfo{person}{Max Jakob}, \bibinfo{person}{Anja Jentzsch},
  \bibinfo{person}{Dimitris Kontokostas}, \bibinfo{person}{Pablo~N. Mendes},
  \bibinfo{person}{Sebastian Hellmann}, \bibinfo{person}{Mohamed Morsey},
  \bibinfo{person}{Patrick van Kleef}, \bibinfo{person}{S{\"o}ren Auer}, {and}
  \bibinfo{person}{Christian Bizer}.} \bibinfo{year}{2014}\natexlab{}.
\newblock \showarticletitle{{DBpedia -- A Large-scale, Multilingual Knowledge
  Base Extracted from Wikipedia}}.
\newblock \bibinfo{journal}{\emph{Semantic Web Journal}} \bibinfo{volume}{6},
  \bibinfo{number}{2} (\bibinfo{year}{2014}), \bibinfo{pages}{167--195}.
\newblock


\bibitem[\protect\citeauthoryear{Ngomo, B{\"{u}}hmann, Unger, Lehmann, and
  Gerber}{Ngomo et~al\mbox{.}}{2013}]%
        {NgomoBULG13a}
\bibfield{author}{\bibinfo{person}{Axel{-}Cyrille~Ngonga Ngomo},
  \bibinfo{person}{Lorenz B{\"{u}}hmann}, \bibinfo{person}{Christina Unger},
  \bibinfo{person}{Jens Lehmann}, {and} \bibinfo{person}{Daniel Gerber}.}
  \bibinfo{year}{2013}\natexlab{}.
\newblock \showarticletitle{{Sorry, I don't Speak SPARQL: Translating SPARQL
  Queries into Natural Language}}. In \bibinfo{booktitle}{\emph{Proceedings of
  the International World Wide Web Conference}}. \bibinfo{pages}{977--988}.
\newblock


\bibitem[\protect\citeauthoryear{Rebele, Suchanek, Hoffart, Biega, Kuzey, and
  Weikum}{Rebele et~al\mbox{.}}{2016}]%
        {RebeleSHBKW16}
\bibfield{author}{\bibinfo{person}{Thomas Rebele}, \bibinfo{person}{Fabian
  Suchanek}, \bibinfo{person}{Johannes Hoffart}, \bibinfo{person}{Joanna
  Biega}, \bibinfo{person}{Erdal Kuzey}, {and} \bibinfo{person}{Gerhard
  Weikum}.} \bibinfo{year}{2016}\natexlab{}.
\newblock \showarticletitle{{YAGO: A Multilingual Knowledge Base from
  Wikipedia, Wordnet, and GeoNames}}. In \bibinfo{booktitle}{\emph{Proceedings
  of the International Semantic Web Conference}}. \bibinfo{pages}{177--185}.
\newblock


\bibitem[\protect\citeauthoryear{Rogers}{Rogers}{2013}]%
        {Rogers:2013}
\bibfield{author}{\bibinfo{person}{Richard Rogers}.}
  \bibinfo{year}{2013}\natexlab{}.
\newblock \bibinfo{booktitle}{\emph{{Digital Methods}}}.
\newblock \bibinfo{publisher}{{MIT Press}}.
\newblock


\bibitem[\protect\citeauthoryear{Rospocher, van Erp, Vossen, Fokkens, Aldabe,
  Rigau, Soroa, Ploeger, and Bogaard}{Rospocher et~al\mbox{.}}{2016}]%
        {RospocherEVFARS16}
\bibfield{author}{\bibinfo{person}{Marco Rospocher}, \bibinfo{person}{Marieke
  van Erp}, \bibinfo{person}{Piek Vossen}, \bibinfo{person}{Antske Fokkens},
  \bibinfo{person}{Itziar Aldabe}, \bibinfo{person}{German Rigau},
  \bibinfo{person}{Aitor Soroa}, \bibinfo{person}{Thomas Ploeger}, {and}
  \bibinfo{person}{Tessel Bogaard}.} \bibinfo{year}{2016}\natexlab{}.
\newblock \showarticletitle{{Building Event-Centric Knowledge Graphs from
  News}}.
\newblock \bibinfo{journal}{\emph{Journal of Web Semantics}}
  \bibinfo{volume}{37-38} (\bibinfo{year}{2016}), \bibinfo{pages}{132--151}.
\newblock


\bibitem[\protect\citeauthoryear{Saquete, Vicedo, Mart{\'\i}nez-Barco, Munoz,
  and Llorens}{Saquete et~al\mbox{.}}{2009}]%
        {SaqueteGMML09}
\bibfield{author}{\bibinfo{person}{Estela Saquete}, \bibinfo{person}{J~Luis
  Vicedo}, \bibinfo{person}{Patricio Mart{\'\i}nez-Barco},
  \bibinfo{person}{Rafael Munoz}, {and} \bibinfo{person}{Hector Llorens}.}
  \bibinfo{year}{2009}\natexlab{}.
\newblock \showarticletitle{{Enhancing {QA} Systems with Complex Temporal
  Question Processing Capabilities}}.
\newblock \bibinfo{journal}{\emph{Journal of Artificial Intelligence Research}}
   \bibinfo{volume}{35} (\bibinfo{year}{2009}), \bibinfo{pages}{775--811}.
\newblock


\bibitem[\protect\citeauthoryear{Shekarpour, Endris, Jaya~Kumar, Lukovnikov,
  Singh, Thakkar, and Lange}{Shekarpour et~al\mbox{.}}{2016}]%
        {shekarpour2016question}
\bibfield{author}{\bibinfo{person}{Saeedeh Shekarpour},
  \bibinfo{person}{Kemele~M Endris}, \bibinfo{person}{Ashwini Jaya~Kumar},
  \bibinfo{person}{Denis Lukovnikov}, \bibinfo{person}{Kuldeep Singh},
  \bibinfo{person}{Harsh Thakkar}, {and} \bibinfo{person}{Christoph Lange}.}
  \bibinfo{year}{2016}\natexlab{}.
\newblock \showarticletitle{{Question Answering on Linked Data: Challenges and
  Future Directions}}. In \bibinfo{booktitle}{\emph{Proceedings of the
  International Conference Companion on World Wide Web}}.
  \bibinfo{pages}{693--698}.
\newblock


\bibitem[\protect\citeauthoryear{Talmor and Berant}{Talmor and Berant}{2018}]%
        {complexwebquestions}
\bibfield{author}{\bibinfo{person}{Alon Talmor} {and} \bibinfo{person}{Jonathan
  Berant}.} \bibinfo{year}{2018}\natexlab{}.
\newblock \showarticletitle{{The Web as a Knowledge-Base for Answering Complex
  Questions}}. In \bibinfo{booktitle}{\emph{Proceedings of the Conference of
  the North American Chapter of the Association for Computational Linguistics:
  Human Language Technologies}}. \bibinfo{pages}{641--651}.
\newblock


\bibitem[\protect\citeauthoryear{Trivedi, Maheshwari, Dubey, and
  Lehmann}{Trivedi et~al\mbox{.}}{2017}]%
        {TrivediMDL17}
\bibfield{author}{\bibinfo{person}{Priyansh Trivedi}, \bibinfo{person}{Gaurav
  Maheshwari}, \bibinfo{person}{Mohnish Dubey}, {and} \bibinfo{person}{Jens
  Lehmann}.} \bibinfo{year}{2017}\natexlab{}.
\newblock \showarticletitle{{{LC-QuAD}: {A} Corpus for Complex Question
  Answering over Knowledge Graphs}}. In \bibinfo{booktitle}{\emph{Proceedings
  of the International Semantic Web Conference}}. \bibinfo{pages}{210--218}.
\newblock


\bibitem[\protect\citeauthoryear{Usbeck, Gusmita, {Axel-Cyrille Ngonga Ngomo},
  and Saleem}{Usbeck et~al\mbox{.}}{2018}]%
        {UsbeckGN018}
\bibfield{author}{\bibinfo{person}{Ricardo Usbeck}, \bibinfo{person}{Ria~Hari
  Gusmita}, \bibinfo{person}{{Axel-Cyrille Ngonga Ngomo}}, {and}
  \bibinfo{person}{Muhammad Saleem}.} \bibinfo{year}{2018}\natexlab{}.
\newblock \showarticletitle{{9th Challenge on Question Answering over Linked
  Data (QALD-9)}}. In \bibinfo{booktitle}{\emph{SemDeep-4 and NLIWOD-4}}.
  \bibinfo{pages}{58--64}.
\newblock


\bibitem[\protect\citeauthoryear{Van~Hage, Malais{\'e}, Segers, Hollink, and
  Schreiber}{Van~Hage et~al\mbox{.}}{2011}]%
        {vanhage:2011}
\bibfield{author}{\bibinfo{person}{Willem~Robert Van~Hage},
  \bibinfo{person}{V{\'e}ronique Malais{\'e}}, \bibinfo{person}{Roxane Segers},
  \bibinfo{person}{Laura Hollink}, {and} \bibinfo{person}{Guus Schreiber}.}
  \bibinfo{year}{2011}\natexlab{}.
\newblock \showarticletitle{{Design and Use of the Simple Event Model (SEM)}}.
\newblock \bibinfo{journal}{\emph{Web Semantics}} \bibinfo{volume}{9},
  \bibinfo{number}{2} (\bibinfo{year}{2011}), \bibinfo{pages}{128--136}.
\newblock


\bibitem[\protect\citeauthoryear{Vrande\v{c}i\'{c}}{Vrande\v{c}i\'{c}}{2012}]%
        {Vrandecic:2012}
\bibfield{author}{\bibinfo{person}{Denny Vrande\v{c}i\'{c}}.}
  \bibinfo{year}{2012}\natexlab{}.
\newblock \showarticletitle{{Wikidata: A New Platform for Collaborative Data
  Collection}}. In \bibinfo{booktitle}{\emph{Proceedings of the International
  Conference on World Wide Web}}. \bibinfo{publisher}{ACM},
  \bibinfo{pages}{1063--1064}.
\newblock


\bibitem[\protect\citeauthoryear{Zafar, Dubey, Lehmann, and Demidova}{Zafar
  et~al\mbox{.}}{2020}]%
        {ZAFAR2020}
\bibfield{author}{\bibinfo{person}{Hamid Zafar}, \bibinfo{person}{Mohnish
  Dubey}, \bibinfo{person}{Jens Lehmann}, {and} \bibinfo{person}{Elena
  Demidova}.} \bibinfo{year}{2020}\natexlab{}.
\newblock \showarticletitle{IQA: Interactive Query Construction in Semantic
  Question Answering Systems}.
\newblock \bibinfo{journal}{\emph{Journal of Web Semantics}}
  \bibinfo{volume}{64} (\bibinfo{year}{2020}), \bibinfo{pages}{100586}.
\newblock
\showISSN{1570-8268}


\end{thebibliography}
\end{document}